\documentclass{article}
\usepackage{spconf,amsmath,graphicx}
\usepackage{algorithm2e}
\usepackage{multirow}
\usepackage{tabularx}
\usepackage{booktabs}

\usepackage{enumitem}
\setlist{nosep, leftmargin=14pt}

\usepackage{mwe} 

\title{Adaptify: A Refined test-time Adaptation Scheme for Frame Classification Consistency in Atrophic Gastritis Videos}
\name{Zinan~Xiong$^{\ast 2}$\thanks{($^{\ast}$Zinan Xiong and Shuijiao Chen are co-first authors.) (Corresponding authors: Yizhe Zhang, Xiaowei Liu)}, Shuijiao~Chen$^{\ast 3}$, Yizhe~Zhang$^{1}$, Yu~Cao$^{2}$, Benyuan~Liu$^{2}$, Xiaowei~Liu$^{3}$
}
\address{\small $^{1}$ School of Computer Science and Engineering, Nanjing University of Science and Technology, Nanjing China\\
\small $^{2}$ 
Miner School of Computer and Information Sciences, University of Massachusetts Lowell, Lowell, MA, 01854 USA\\
\small $^{3}$ Department of Gastroenterology, Xiangya Hospital of Central South University, Changsha, Hunan 410008 China
}
\begin{document}
%
\maketitle
\begin{abstract}
Atrophic gastritis is a significant risk factor for developing gastric cancer. The incorporation of machine learning algorithms can efficiently elevate the possibility of accurately detecting atrophic gastritis. Nevertheless, when the trained model is applied in real-life circumstances, its output is often not consistently reliable. In this paper, we propose Adaptify, an adaptation scheme in which the model assimilates knowledge from its own classification decisions. Our proposed approach includes keeping the primary model constant, while simultaneously running and updating the auxiliary model. By integrating the knowledge gleaned by the auxiliary model into the primary model and merging their outputs, we have observed a notable improvement in output stability and consistency compared to relying solely on either the main model or the auxiliary model.
\end{abstract}
\section{Introduction}
\label{sec:intro}

Atrophic gastritis is a chronic inflammatory condition that affects the lining of the stomach.
The diagnosis of atrophic gastritis involves a combination of clinical symptoms, laboratory tests, and imaging studies. 
Machine learning (ML) has gained prominence in recent years as a powerful tool in healthcare research, diagnosis, and treatment \cite{9874457}, \cite{zhang2022automated}, \cite{10098014}, \cite{BAO2023126991}, \cite{wang2023dental}, \cite{ZHANG2023100393}, \cite{bao2023learning}. 
While they have made significant strides in image classification within video streams, challenges persist in achieving consistent results.  


Temporal consistency is vital not only in segmentation but also in image classification tasks. In cases such as atrophic gastritis, where using segmentation masks or bounding boxes is challenging, whole image classification methods are more commonly used. Thus, achieving temporal consistency in video image classification requires innovative approaches to maintain stable predictions across frames. 

In this paper, we introduces an unsupervised and efficient method Adaptify, inspired by AuxAdapt \cite{zhang2022auxadapt}, to address the challenges of consistent image frame classification across video streams. Our proposed method utilizes two networks: a main network, with fixed parameters during test time, and an auxiliary network that undergoes continuous parameter updates. Furthermore, to guarantee smooth transitions between frames, our method incorporates classification outputs from multiple preceding frames by assigning them specific weights. These weighted outputs are then combined with the classification score of the current frame. As a result, the final score incorporates information from both preceding frames and the current frame, yielding a more robust and consistent classification result. 

In order to comprehensively assess the effectiveness and adaptability of our model during test time, we conducted comprehensive experiments using our image dataset and video dataset. 
The results demonstrate that all combinations outperform individual models, significantly reducing both false positive and false negative predictions.

\section{Related Work}

\textbf{Image Classification:} Image classification is an important problem in computer vision. Over the years, many deep learning architectures have been proposed for this task, including AlexNet \cite{krizhevsky2017imagenet} and ResNet \cite{he2016deep}. 
The advancement of deep learning models has opened up new opportunities for medical image analysis \cite{yadav2019deep}, \cite{zhang2023deep}. Unlike traditional image classification algorithms, deep learning models can automatically extract and learn features from the input data without the need for hand-crafted feature selection. 


\textbf{Temporal Consistency:} Temporal consistency is a crucial objective in tasks like semantic segmentation. When algorithms are applied to video tasks, it becomes essential to ensure that the semantic labels of the same object in adjacent frames are consistent. To address this challenge, various techniques have been developed \cite{zhang2024testfit}, \cite{hur2016joint}. 
Cheng \emph{et al.} \cite{cheng2017segflow} proposed an end-to-end network that addresses the challenge of simultaneous pixel-wise object segmentation and optical flow prediction in videos. Ding \emph{et al.} \cite{ding2020every} introduced a similar design in which they jointly trained segmentation and optical flow models while incorporating a novel loss function. 

\begin{figure}[t]
    \centering    \includegraphics[width=0.9\columnwidth]{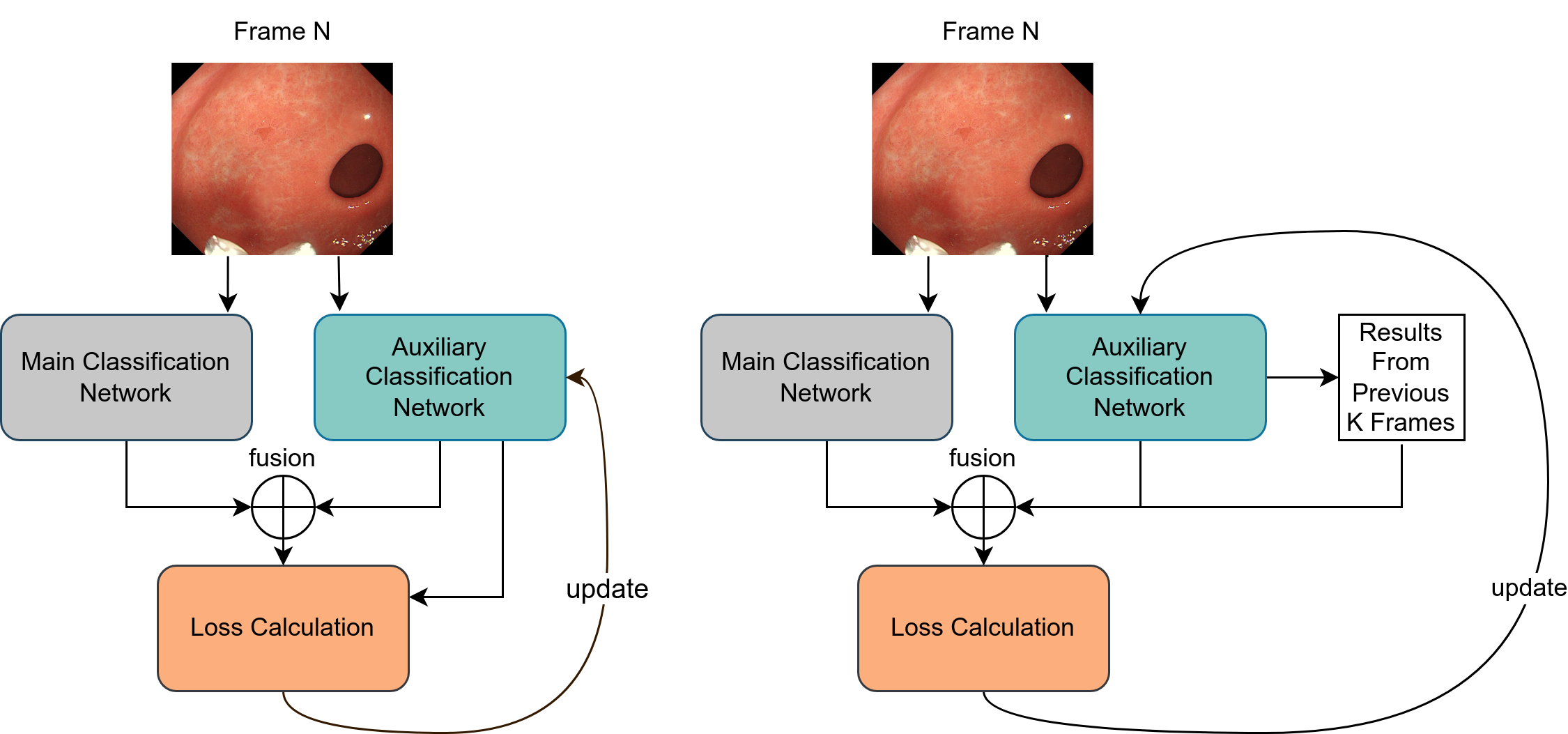}
    \caption{Left: Adaptation considering 1 frame. Right: Adaptation considering K frames.}
    \label{fig:flowchart}
\end{figure}

\textbf{Test-Time Adaptation: }Test-time adaptation refers to the process of adjusting the model's predictions or parameters during the testing phase. By employing test-time adaptation techniques, models can effectively adapt to novel data distributions and enhance their performance on previously unseen examples. Wang \emph{et al.} \cite{wang2020tent} presented a novel approach for fully test-time adaptation by employing test entropy minimization. Yi \emph{et al.} \cite{yi2023temporal} introduced Temporal Coherent Test-time Optimization (TeCo), aimed at harnessing spatiotemporal insights for robust video classification during test-time optimization. Zhang \emph{et al.} \cite{zhang2022auxadapt} introduced the AuxAdapt scheme, which leverages a small auxiliary segmentation network to improve the performance of the main segmentation network in video streams.
\vspace{-10pt}
\section{Method}
Due to the sensitivity characteristics of deep learning algorithms to small changes, directly applying the algorithm to video sequences would yield inconsistent results. 

In this section, we present the details of our proposed method as illustruated in Fig. \ref{fig:flowchart}, which focuses on enhancing temporal consistency through the integration of a small auxiliary network.

Our proposed model draws inspiration from the concept of AuxAdapt \cite{zhang2022auxadapt}, making adaptations to suit the requirements of the image classification task. By fixing the parameters of the main model, we ensure a stable and reliable output. Simultaneously, a small auxiliary model continuously learns from the current frame and preceding frames, and updating its parameters during test time. This auxiliary model serves as a refinement model, working in conjunction with the main model's output to adjust and enhance the final classification result, thereby promoting greater output consistency. 

\subsection{AuxAdapt}


To ensure a reliable and uniform outcome, it becomes imperative and advantageous for the model to acquire knowledge from its own decision-making process. AuxAdapt incorporates a compact auxiliary network that assists in adapting the model at test-time. To determine the extent to which information from previous frames should be leveraged, a momentum scheme is implemented. By incorporating this scheme, the model gains the ability to intelligently gauge the relevance and impact of past frames, thus enabling a more informed decision-making process.

\subsection{Adaptify}
In contrast to the instance segmentation task, which assigns probabilities to each pixel on the feature map with respect to specific object classes, image classification aims to predict a single probability score for the entire image. This characteristic makes image classification more susceptible to small fluctuations in the input data, thus requires robust and stable feature representations to ensure consistent predictions.

Unlike AuxAdapt, which only utilizes the predicted feature map from latest two frames, Adaptify takes a different approach by leveraging the classification results obtained from several previous frames.

As demonstrated in Algorithm \ref{alg:two}, the initialization phase involves creating a buffer $B$ of size $K$. The video sequence is then sequentially passed through both the main network and the auxiliary network frame by frame. During the initial stages of the algorithm, when the auxiliary network generates its output, the outcome is first appended to buffer $B$. Subsequently, the probability vector originating from the main network is augmented by the summation of the probability vector generated by the auxiliary network. 
When the number of elements within buffer $B$ reaches the limit $K$, a mechanism is triggered to maintain a rolling window of the most recent results, ensuring that the buffer constantly stores the latest $K$ outcomes.

Subsequently, the algorithm combines the main network's output with the cumulative sum of elements within buffer $B$. A cross-entropy loss is then computed as follows: 

\begin{equation}\label{loss_2}
    \mathcal{L}\left(y_t^{\text {aux }}, y_t^{\mathrm{cls}}\right)=\mathcal{L}_{CE}{\left(y_t^{\operatorname{aux}}, y_t^{\mathrm{cls}}\right)}
\end{equation}
where $\mathcal{L}_{CE}$ represents the cross-entropy loss, $y_t^{aux}$ is the output from auxiliary model, $y_t^{cls}$ is the combined result of two models.

The parameters of the auxiliary network are updated using the following rule:

\begin{equation}\label{parameter_update_2}
    \begin{aligned}
& \Delta \theta_t^{\text {aux }}=\gamma \Delta \theta_{t-1}^{\text {aux }}+\lambda \nabla_{\theta^{\text {aux }}} \mathcal{L}\left(y_t^{\text {aux }}, y_t^{\text {cls }}\right) \\
& \theta_t^{\text {aux }}=\theta_{t-1}^{\text {aux }}+\Delta \theta_t^{\text {aux }}
\end{aligned}
\end{equation}
where $\gamma$ is the learning rate, $\lambda$ is the momentum coefficient, and $\theta^{aux}$ is the parameters of the auxiliary model.


Employing an argmax function, the algorithm identifies the index corresponding to the highest value in the final result vector, ultimately yielding the definitive classification outcome.

\begin{figure}[t!]
    \centering    \includegraphics[width=0.9\columnwidth]{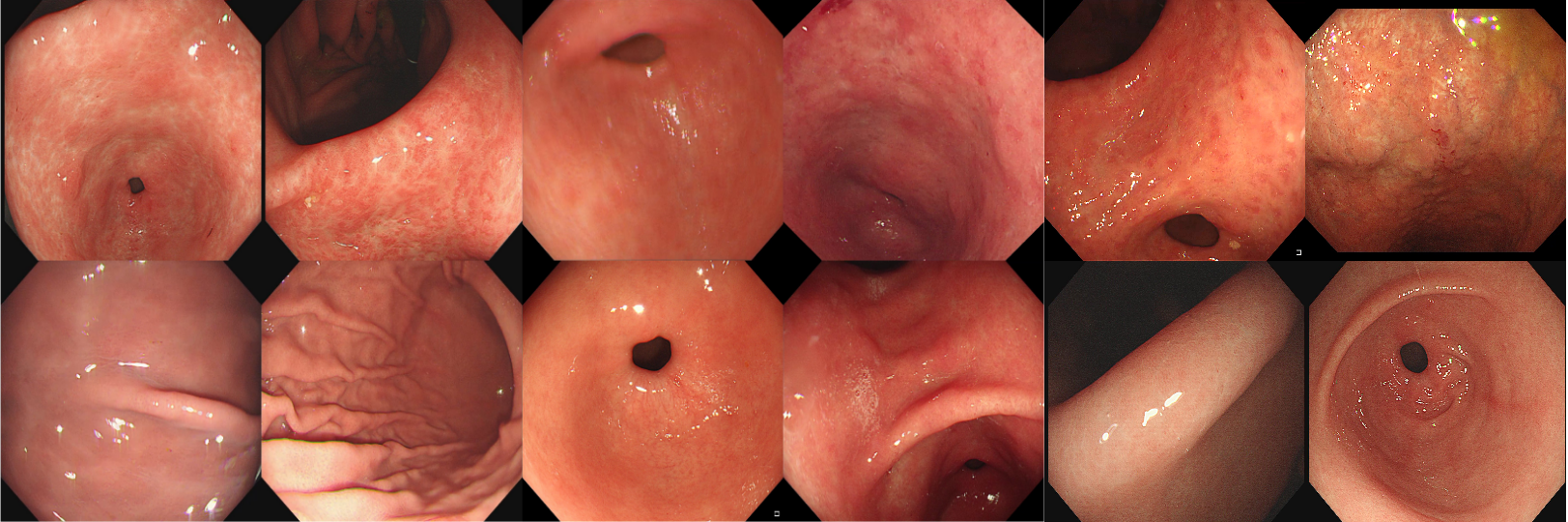}
    \caption{Image exampled from training dataset. Top: atrophic gastritis images; Bottom: healthy images}
    \label{fig:dataset}
\end{figure}


\RestyleAlgo{ruled}
\begin{algorithm}[t!]
\caption{Adaptify}\label{alg:two}
\SetKwInput{KwData}{Input}
\SetKwInput{KwResult}{Output}
\KwData{frames: $x_1, x_2, \ldots, x_T ;$ buffer $B$ of size $K$}
\KwResult{$y_1^{\text {cls}}, y_2^{\text {cls}}, \ldots, y_T^{\text {cls}} ;$}
Load trained MainNet $f^{\text {main}}$, which will be frozen;\\
Load trained AuxNet $f^{\text {aux}}$ as $f_1^{\text {aux}}$;\\

Initialize sequence number $t=1 ;$ \\

\While{$t < T$}{
$y_t^{\text {main }}=f^{\text {main }}\left(x_t\right),{ } y_t^{\text {aux }}=f_t^{\text {aux }}\left(x_t\right) ;$ \\
\eIf{$B.size < K$}
{$y_t^{\text {cls }}={\operatorname{argmax}}\  (\alpha y_t^{\text {main }}+\beta y_t^{\text {aux }}) ;$\\
$B.append(y_t^{\text {aux }})$}{$B.pop(0)$;\\
$B.append(y_t^{\text {aux }})$;\\
$y_t^{\text {cls }}={\operatorname{argmax}}\  (\alpha y_t^{\text {main }}+\beta \sum_{i=1}^{K}b_i) ;$} 
Compute loss: $\mathcal{L}\left(y_t^{\text {aux }}, y_t^{\text {cls }}\right)$ using Eq. (3); 

Update $f_t^{\text {aux }}$ using Eq. (4), which gives $f_{t+1}^{\text {aux }} ;$ 

$t \leftarrow t+1 ;$ 
}
\end{algorithm}

To regulate the significance of the current frame versus prior frames, we introduce two hyperparameters, $\alpha$ and $\beta$, serving the purpose of modulating the weighting attributed to the present frame and the stored frames within the buffer.
Through carefully selecting $\alpha$, $\beta$, and $K$, the model attains heightened flexibility in calibrating the relative importance of the current frame and the historical frames within the buffer.

By incorporating classification results from past frames, our method introduces temporal context into the classification process. This temporal context provides valuable insights into the object's evolution over time, allowing the model to make more informed decisions by considering the historical class information.



%

\section{Experiments}
\textbf{Dataset: }The training dataset is composed of a collection of 3,397 images with atrophic gastritis and 2,739 images depicting healthy instances. The validation set encompasses an equal number of 200 images for each respective category. These images have been derived from videos captured by medical professionals during gastroscopy procedures. A few sample images from our training dataset is shown in Figure \ref{fig:dataset}.

\textbf{Models:} ResNet-50 is selected as the main model. Additionally, we incorporate smaller auxiliary networks including ResNet-18, MobileNet-V2, and EfficientNet-B3 \cite{tan2019efficientnet}. These auxiliary networks offer reduced parameter complexity while still contributing valuable insights to the overall framework. Furthermore, various combinations of coefficients $\alpha$ and $\beta$ were also explored to adjust the significance attributed to the current frame and previous frames.

\begin{table}[]
\caption{validation accuracy of several baseline models}
\label{tab:baseline}
\resizebox{\columnwidth}{!}{\begin{tabular}{|l|l|l|l|l|}
\hline
Model                                                                               & ResNet-50                   & ResNet-18                   & EfficientNet-B3             & MobileNet-V2                \\ \hline
\multicolumn{1}{|c|}{\begin{tabular}[c]{@{}c@{}}Validation\\ Accuracy\end{tabular}} & \multicolumn{1}{c|}{0.8250} & \multicolumn{1}{c|}{0.7975} & \multicolumn{1}{c|}{0.8775} & \multicolumn{1}{c|}{0.7575} \\ \hline
\end{tabular}}
\end{table}
\vspace{-15pt}

\begin{table}[h]
\caption{Evaluation results of ResNet-50 and ResNet-18}
\label{tab:resnet18}
\resizebox{\columnwidth}{!}{\begin{tabular}{|l|l|l|l|l|}
\hline
\begin{tabular}[c]{@{}l@{}}model\\ combination\end{tabular} & \begin{tabular}[c]{@{}l@{}}frames\\ considered\end{tabular} & alpha                                             & beta                                                  & \begin{tabular}[c]{@{}l@{}}validation\\ accuracy\end{tabular}    \\ \hline
\multicolumn{1}{|c|}{ResNet-50 + ResNet-18}                 & 1                                                           & \begin{tabular}[c]{@{}l@{}}1\\ 1\\ 1\end{tabular} & \begin{tabular}[c]{@{}l@{}}1\\ 0.8\\ 0.5\end{tabular} & \begin{tabular}[c]{@{}l@{}}0.8350\\ 0.8175\\ 0.7800\end{tabular} \\ \hline
ResNet-50 + ResNet-18                                       & 3                                                           & \begin{tabular}[c]{@{}l@{}}1\\ 1\\ 1\end{tabular} & \begin{tabular}[c]{@{}l@{}}1\\ 0.8\\ 0.5\end{tabular} & \begin{tabular}[c]{@{}l@{}}0.8225\\ 0.7875\\ 0.7350\end{tabular} \\ \hline
ResNet-50 + ResNet-18                                       & 4                                                           & \begin{tabular}[c]{@{}l@{}}1\\ 1\\ 1\end{tabular} & \begin{tabular}[c]{@{}l@{}}1\\ 0.8\\ 0.5\end{tabular} & \begin{tabular}[c]{@{}l@{}}0.8325\\ 0.7750\\ 0.7925\end{tabular} \\ \hline
\end{tabular}}
\end{table}

\vspace{-15pt}

\begin{table}[h]
\caption{Evaluation results of ResNet-50 and MobileNet-V2}
\label{tab:mobilenetv2}
\resizebox{\columnwidth}{!}{\begin{tabular}{|l|l|l|l|l|}
\hline
\begin{tabular}[c]{@{}l@{}}model\\ combination\end{tabular} & \begin{tabular}[c]{@{}l@{}}frames\\ considered\end{tabular} & alpha                                             & beta                                                  & \begin{tabular}[c]{@{}l@{}}validation\\ accuracy\end{tabular}    \\ \hline
\multicolumn{1}{|c|}{ResNet-50 + MobileNet-V2}                 & 1                                                           & \begin{tabular}[c]{@{}l@{}}1\\ 1\\ 1\end{tabular} & \begin{tabular}[c]{@{}l@{}}1\\ 0.8\\ 0.5\end{tabular} & \begin{tabular}[c]{@{}l@{}}0.7725\\ 0.8000\\ 0.8575\end{tabular} \\ \hline
ResNet-50 + MobileNet-V2                                       & 3                                                           & \begin{tabular}[c]{@{}l@{}}1\\ 1\\ 1\end{tabular} & \begin{tabular}[c]{@{}l@{}}1\\ 0.8\\ 0.5\end{tabular} & \begin{tabular}[c]{@{}l@{}}0.8433\\ 0.8250\\ 0.7775\end{tabular} \\ \hline
ResNet-50 + MobileNet-V2                                       & 4                                                           & \begin{tabular}[c]{@{}l@{}}1\\ 1\\ 1\end{tabular} & \begin{tabular}[c]{@{}l@{}}1\\ 0.8\\ 0.5\end{tabular} & \begin{tabular}[c]{@{}l@{}}0.8025\\ 0.8100\\ 0.8000\end{tabular} \\ \hline
\end{tabular}}
\end{table}

\begin{table}[h]
\caption{Evaluation results of ResNet-50 and EfficientNet-B3}
\label{tab:efficientnetb3}
\resizebox{\columnwidth}{!}{\begin{tabular}{|l|l|l|l|l|}
\hline
\begin{tabular}[c]{@{}l@{}}model\\ combination\end{tabular} & \begin{tabular}[c]{@{}l@{}}frames\\ considered\end{tabular} & alpha                                             & beta                                                  & \begin{tabular}[c]{@{}l@{}}validation\\ accuracy\end{tabular}    \\ \hline
\multicolumn{1}{|c|}{ResNet-50 + EfficientNet-B3}                 & 1                                                           & \begin{tabular}[c]{@{}l@{}}1\\ 1\\ 1\end{tabular} & \begin{tabular}[c]{@{}l@{}}1\\ 0.8\\ 0.5\end{tabular} & \begin{tabular}[c]{@{}l@{}}0.7850\\ 0.8400\\ 0.8075\end{tabular} \\ \hline
ResNet-50 + EfficientNet-B3                                       & 3                                                           & \begin{tabular}[c]{@{}l@{}}1\\ 1\\ 1\end{tabular} & \begin{tabular}[c]{@{}l@{}}1\\ 0.8\\ 0.5\end{tabular} & \begin{tabular}[c]{@{}l@{}}0.8225\\ 0.7975\\ 0.8275\end{tabular} \\ \hline
ResNet-50 + EfficientNet-B3                                       & 4                                                           & \begin{tabular}[c]{@{}l@{}}1\\ 1\\ 1\end{tabular} & \begin{tabular}[c]{@{}l@{}}1\\ 0.8\\ 0.5\end{tabular} & \begin{tabular}[c]{@{}l@{}}0.7575\\ 0.7650\\ 0.7950\end{tabular} \\ \hline
\end{tabular}}
\end{table}

\begin{figure}[hb!]
    \centering    \includegraphics[width=\columnwidth]{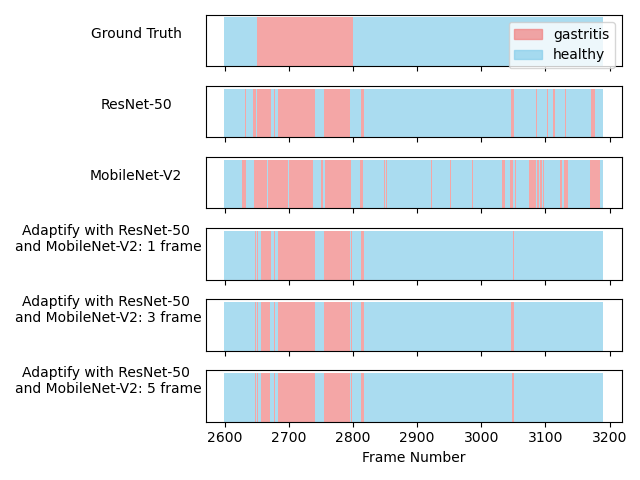}
    \caption{Performance Evaluation on a single video using ResNet-50 and MobileNet-V2. The first one represents the ground truth. The horizontal axis represents the frame number.}
    \label{fig:videoresult_mobilenet}
\end{figure}

\begin{figure}[h!]
    \centering    \includegraphics[width=\columnwidth]{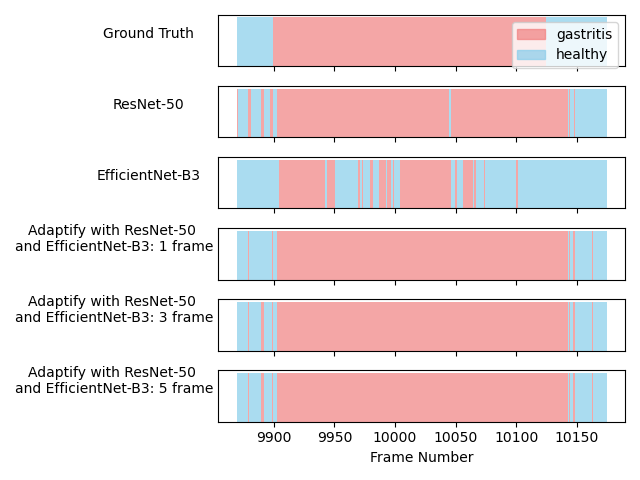}
    \caption{Performance Evaluation on a single video using ResNet-50 and EfficientNet-B3. The horizontal axis represents the frame number.}
    \label{fig:videoresult}
\end{figure}

\textbf{Evaluation metric: }After completing the training phase, we establish a baseline for fair comparison by initially running all models exclusively on the dataset, and the outcomes are presented in Table \ref{tab:baseline}. 
Before subjecting the validation images, the model undergoes an adaptation process utilizing a continuous sequence of 200 frames that immediately precede these images, from $t-200$ to $t-1$, establishing a coherent temporal context. 
Subsequent to test-time adaptation, the model proceeds to predict outcomes for the validation images. 

Table \ref{tab:resnet18} reveals that when ResNet-50 serves as the main model and ResNet-18 acts as the auxiliary model in scenarios considering 1 frame, 3 frames, and 5 frames, the combination with alpha set to 1 and beta set to 1 yields the highest accuracy scores: 0.8350, 0.8225, and 0.8325, respectively. These results surpass the accuracy achieved when using ResNet-50 or ResNet-18 individually. 
In Table \ref{tab:mobilenetv2}, employing MobileNet-V2 as the auxiliary model, Adaptify achieves its highest accuracy when considering 1 frame and setting beta to 0.5, resulting in an accuracy of 0.8575, which is also higher than using these two models individually. 
From Table \ref{tab:efficientnetb3}, it's worth noting that although the highest accuracy achieved with this combination is 0.8400, which is lower than using EfficientNet-B3 alone (0.8775), its performance significantly improves when applied to the video dataset, as we will demonstrate later.

We then conducted evaluations on these models using a video dataset. As depicted in Fig.\ref{fig:videoresult_mobilenet} and Fig.\ref{fig:videoresult}, we carried out five distinct experiments. The initial experiment exclusively employed the ResNet-50 model, followed by a second experiment utilizing the auxiliary model. Subsequently, the remaining three experiments featured our innovative Adaptify model, each considering a different number of frames for analysis.
The results yield a noteworthy observation: the application of Adaptify to the ResNet-50 model effectively smoothed its output while maintaining stability across all experiments. The results in Fig. 4 indicate a notable reduction in the number of false positive predictions (frames before 2,650 and after 2,800) and false negative predictions. Likewise, in Fig. 5, we observe a substantial decrease in false positive predictions before frame 9,900 and false negative predictions between frame 9,900 and frame 10,100, leading to a considerably more consistent output. An important aspect to highlight is the apparent disparity between Table \ref{tab:baseline} and Table \ref{tab:efficientnetb3}. It suggests that the combination of ResNet-50 and EfficientNet-B3 may yield inferior results when compared to using EfficientNet-B3 alone. However, an intriguing revelation emerges from Fig. \ref{fig:videoresult}, where it becomes evident that the Adaptify combination outperforms others by effectively mitigating the occurrence of false negative spikes in predictions. 



\section{Conclusion}
Within the context of this paper, we put forth Adaptify, an unsupervised online test-time adaptation classifier model. The primary objective of this model is to bolster the temporal coherence inherent to frame classifications within video sequences. By harnessing the capabilities of a compact auxiliary network and ingeniously considering a range of preceding frames for the current frame, our proposed methodology exhibits a remarkable capacity for elevating the quality of frame classifications.

The experimental results underscore the efficacy of our approach. Through the utilization of the auxiliary network and the strategic incorporation of temporal context, Adaptify achieves a considerable augmentation in the quality of frame classifications within video streams. 

\bibliographystyle{IEEEbib}
\bibliography{strings,refs}

\begin{thebibliography}{10}

\bibitem{9874457}
Zinan Xiong, Chenxi Wang, Ying Li, Yan Luo, and Yu~Cao,
\newblock ``Swin-pose: Swin transformer based human pose estimation,''
\newblock in {\em 2022 IEEE 5th International Conference on Multimedia Information Processing and Retrieval (MIPR)}, 2022, pp. 228--233.

\bibitem{zhang2022automated}
Chenxi Zhang, Zinan Xiong, Shuijiao Chen, Alex Ding, Yu~Cao, Benyuan Liu, and Xiaowei Liu,
\newblock ``Automated disease detection in gastroscopy videos using convolutional neural networks,''
\newblock {\em Frontiers in Medicine}, vol. 9, pp. 846024, 2022.

\bibitem{10098014}
Zinan Xiong, Qilei Chen, Chenxi Zhang, Yu~Cao, Benyuan Liu, Yu~Wu, Yu~Peng, and Xiaowei Liu,
\newblock ``Deep learning assisted mouth-esophagus passage time estimation during gastroscopy,''
\newblock in {\em 2022 IEEE 34th International Conference on Tools with Artificial Intelligence (ICTAI)}, 2022, pp. 1105--1111.

\bibitem{BAO2023126991}
Yajie Bao, Tianwei Xing, and Xun Chen,
\newblock ``Confidence-based interactable neural-symbolic visual question answering,''
\newblock {\em Neurocomputing}, p. 126991, 2023.

\bibitem{wang2023dental}
Xizhe Wang, Jing Guo, Peng Zhang, Qilei Chen, Zhang Zhang, Yu~Cao, Xinwen Fu, and Benyuan Liu,
\newblock ``A deep learning framework with pruning roi proposal for dental caries detection in panoramic x-ray images,''
\newblock in {\em Neural Information Processing: 30th International Conference, ICONIP 2023}. Springer, 2023.

\bibitem{ZHANG2023100393}
Zhang Zhang, Qilei Chen, Shuijiao Chen, Xiaowei Liu, Yu~Cao, Benyuan Liu, and Honggang Zhang,
\newblock ``Automatic disease detection in endoscopy with light weight transformer,''
\newblock {\em Smart Health}, vol. 28, pp. 100393, 2023.

\bibitem{bao2023learning}
Yajie Bao, Kimberly~J Chan, Ali Mesbah, and Javad~Mohammadpour Velni,
\newblock ``Learning-based adaptive-scenario-tree model predictive control with improved probabilistic safety using robust bayesian neural networks,''
\newblock {\em International Journal of Robust and Nonlinear Control}, vol. 33, no. 5, pp. 3312--3333, 2023.

\bibitem{zhang2022auxadapt}
Yizhe Zhang, Shubhankar Borse, Hong Cai, and Fatih Porikli,
\newblock ``Auxadapt: Stable and efficient test-time adaptation for temporally consistent video semantic segmentation,''
\newblock in {\em Proceedings of the IEEE/CVF Winter Conference on Applications of Computer Vision}, 2022, pp. 2339--2348.

\bibitem{krizhevsky2017imagenet}
Alex Krizhevsky, Ilya Sutskever, and Geoffrey~E Hinton,
\newblock ``Imagenet classification with deep convolutional neural networks,''
\newblock {\em Communications of the ACM}, vol. 60, no. 6, pp. 84--90, 2017.

\bibitem{he2016deep}
Kaiming He, Xiangyu Zhang, Shaoqing Ren, and Jian Sun,
\newblock ``Deep residual learning for image recognition,''
\newblock in {\em Proceedings of the IEEE conference on computer vision and pattern recognition}, 2016, pp. 770--778.

\bibitem{yadav2019deep}
Samir~S Yadav and Shivajirao~M Jadhav,
\newblock ``Deep convolutional neural network based medical image classification for disease diagnosis,''
\newblock {\em Journal of Big data}, vol. 6, no. 1, pp. 1--18, 2019.

\bibitem{zhang2023deep}
Chenxi Zhang, Alexander Ding, Zhehong Fu, Jing Ni, Qilei Chen, Zinan Xiong, Benyuan Liu, Yu~Cao, Shuijiao Chen, and Xiaowei Liu,
\newblock ``Deep learning for gastric location classification: An analysis of location boundaries and improvements through attention and contrastive learning,''
\newblock {\em Smart Health}, vol. 28, pp. 100394, 2023.

\bibitem{zhang2024testfit}
Yizhe Zhang, Tao Zhou, Yuhui Tao, Shuo Wang, Ye~Wu, Benyuan Liu, Pengfei Gu, Qiang Chen, and Danny~Z Chen,
\newblock ``Testfit: A plug-and-play one-pass test time method for medical image segmentation,''
\newblock {\em Medical Image Analysis}, vol. 92, pp. 103069, 2024.

\bibitem{hur2016joint}
Junhwa Hur and Stefan Roth,
\newblock ``Joint optical flow and temporally consistent semantic segmentation,''
\newblock in {\em Computer Vision--ECCV 2016 Workshops: Amsterdam, The Netherlands, October 8-10 and 15-16, 2016, Proceedings, Part I 14}. Springer, 2016, pp. 163--177.

\bibitem{cheng2017segflow}
Jingchun Cheng, Yi-Hsuan Tsai, Shengjin Wang, and Ming-Hsuan Yang,
\newblock ``Segflow: Joint learning for video object segmentation and optical flow,''
\newblock in {\em Proceedings of the IEEE international conference on computer vision}, 2017, pp. 686--695.

\bibitem{ding2020every}
Mingyu Ding, Zhe Wang, Bolei Zhou, Jianping Shi, Zhiwu Lu, and Ping Luo,
\newblock ``Every frame counts: Joint learning of video segmentation and optical flow,''
\newblock in {\em Proceedings of the AAAI Conference on Artificial Intelligence}, 2020, vol.~34, pp. 10713--10720.

\bibitem{wang2020tent}
Dequan Wang, Evan Shelhamer, Shaoteng Liu, Bruno Olshausen, and Trevor Darrell,
\newblock ``Tent: Fully test-time adaptation by entropy minimization,''
\newblock {\em arXiv preprint arXiv:2006.10726}, 2020.

\bibitem{yi2023temporal}
Chenyu Yi, Siyuan Yang, Yufei Wang, Haoliang Li, Yap-Peng Tan, and Alex~C. Kot,
\newblock ``Temporal coherent test-time optimization for robust video classification,'' 2023.

\bibitem{tan2019efficientnet}
Mingxing Tan and Quoc Le,
\newblock ``Efficientnet: Rethinking model scaling for convolutional neural networks,''
\newblock in {\em International conference on machine learning}. PMLR, 2019, pp. 6105--6114.

\end{thebibliography}

\end{document}